# Determinantal Clustering Process - A Nonparametric Bayesian Approach to Kernel Based Semi-Supervised Clustering

**Amar Shah**
Department of Engineering
University of Cambridge
`as793@cam.ac.uk`

**Zoubin Ghahramani**
Department of Engineering
University of Cambridge
`zoubin@eng.cam.ac.uk`

## Abstract

Semi-supervised clustering is the task of clustering data points into clusters where only a fraction of the points are labelled. The true number of clusters in the data is often unknown and most models require this parameter as an input. Dirichlet process mixture models are appealing as they can infer the number of clusters from the data. However, these models do not deal with high dimensional data well and can encounter difficulties in inference. We present a novel nonparameteric Bayesian method to cluster data points without the need to prespecify the number of clusters or to model complicated densities from which data points are assumed to be generated from. The key insight is to use determinants of submatrices of a kernel matrix as a measure of how close together a set of points are. We explore some theoretical properties of the model and derive a natural Gibbs based algorithm with MCMC hyperparameter learning. We test the model on various synthetic and real world data sets.

## 1 INTRODUCTION

Finding clusters amongst data points has been a key idea addressed by many researchers in machine learning, statistics and signal processing. In practice it is often the case that copious amounts of data can be easily collected, but subsequent labelling of the data is expensive and slow to obtain. *Semi-supervised learning* algorithms aim to utilise information from both labelled and unlabelled data to inform choices about how best to partition data points or where decision boundaries lie.

A natural approach for a Bayesian practitioner would be to consider a *generative model* of the data. This involves explicit modelling of the density which produced the observations and averaging over possible clusterings of the unlabelled training data. Next any parameters of the density model can be integrated out to produce a predictive clustering of unseen test data.

The choice of density model is integral and highly influential on the results of the training. A nonparametric Bayesian model is attractive in that it can incorporate an unbounded number of parameters and is able, in theory, to learn the correct density model e.g. Dirichlet Process Mixture Model [Escobar and West, 1995]. However such an approach can be expensive as typically, when clustering, we are not interested in the density itself whilst much effort is spent in learning it. Adams and Ghahramani [2009] propose a fully-Bayesian generative approach to semi-supervised classification which avoids the need to model complex density functions. Nonetheless this model does require prior specification of the number of classes and the training of a Gaussian process per class.

Discriminative models tend to be more popular for Bayesian semi-supervised learning [Zhu, 2005]. Lawrence and Jordan [2005] construct a nonparametric Bayesian model for binary semi-supervised classification, which is extended to the multi-class case by Rogers and Girolami [2007]. Similar Gaussian process based discriminative models that exploit graph-based information are suggested by Chu et al. [2007] and Sindhwani et al. [2007]. All of these examples also require knowledge of the total number of classes that the data is divided into.

In this work we present a novel discriminative nonparametric Bayesian method for clustering points using a kernel matrix determinant based measure of similarity between data points. It is nonparametric in that prior mass is assigned to all possible partitions of the data. This method is highly appropriate in the case where a generative model is computationally prohibitively expensive to train but where a kernel between pairs of data points can be easily computed e.g. in high dimensional data which cannot be adequately represented on a low dimensional manifold. Thus we bring together some of the most attracive properties of discriminative kernel methods (removing the need to model the input observations) and Bayesian nonparametrics (the ability to infer the number of clusters and kernel hyperparameters).

Our model makes use of a popular determinant based likelihood model called the *determinantal point process* (DPP) [Kulesza and Taskar, 2013]. This is a prior on the set of all subsets of a data set which places higher mass on subsets which contain diverse elements. DPPs arise in classical theory such as random matrix theory [Mehta and Gaudin, 1960, Ginibre, 1965] and quantum physics [Macchi, 1975]. They have more recently been used for human pose estimation, search diversification and document summarization [Kulesza and Taskar, 2010, 2011a,b].

In Section 2 we introduce the determinantal clustering process likelihood and discuss some of its properites. In Section 3 we explain how such a model can be applied to semi supervised clustering problems and develop a Gibbs based inference scheme with MCMC hyperparameter updates. We consider related research in Section 4 and describe the novel features of the DCP amongst other algorithms. Experimental performance of the model is outlined in Section 5 and finally conclusions are discussed in Section 6.

## 2 THE DETERMINANTAL CLUSTERING MODEL

Consider data points $X = \{x_n\}_{n=1}^N$ which live in a space $\mathcal{X}$. We assume that $x_i \neq x_j$ whenever $i \neq j$ without loss of generality (since we can assign any pair of points which violate this condition to the same cluster). In this work we typically consider the case $\mathcal{X} = \mathbb{R}^D$, but the ideas can easily be extended to more general spaces. Suppose further that we are given a positive definite kernel function, $k$, which depends on hyperparameters belonging to a space $\Theta$, such that $k : \mathcal{X} \times \mathcal{X} \times \Theta \to \mathbb{R}$.

For a given set of hyperparameters, $\theta \in \Theta$ and ordered subsets $A, B \subseteq X$ of sizes $N_A$, $N_B$, define the $N_A \times N_B$ Gram matrix $K_{A,B}^\theta$ as

$$K_{A,B}^\theta(i,j) = k(x_{a_i}, x_{b_j}, \theta), \qquad (1)$$

where $x_{a_i}$ is the $i^{th}$ element in $A$ and $x_{b_j}$ is the $j^{th}$ element in $B$. For notational convenience we write $K_{A,A}^\theta$ as $K_A^\theta$.

Let $\mathbb{S}$ be the set of all partitions of $X$. Hence an element $\mathcal{S} \in \mathbb{S}$ is a set of subsets of $X$ such that for any $S, S' \in \mathcal{S}$, $S \cap S' = \emptyset$ and $\bigcup_{S \in \mathcal{S}} S = X$. We introduce the notion of a *determinantal clustering process* (DCP) for a given kernel function $k$ and hyperparameters $\theta \in \Theta$, defined as a probability measure on the set $\mathbb{S}$ with density function

$$p(\mathcal{S} \in \mathbb{S}|\theta) \propto \prod_{S \in \mathcal{S}} \det\left(K_S^\theta\right)^{-1}. \qquad (2)$$

Note that whilst DPPs have been extensively used as priors over subsets to encourage diversity [Kulesza and Taskar, 2013], we use the *inverse* of the determinant as a measure of similarity amongst points in a cluster which has very different properties which we discuss in the next section.

### 2.1 PROPERTIES OF THE DCP

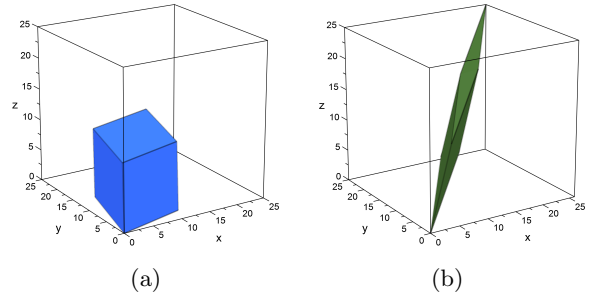

Figure 1: In (a) the volume represents the determinant of $\begin{pmatrix} 10 & 1 & 2 \\ 1 & 10 & 1 \\ 2 & 1 & 10 \end{pmatrix}$ whilst the in (b) the volume represents the determinant of $\begin{pmatrix} 10 & 8 & 7 \\ 8 & 10 & 8 \\ 7 & 8 & 10 \end{pmatrix}$. The more similar the rows of the matrix are, the smaller the determinant is.

Note that for any $S \in \mathcal{S} \in \mathbb{S}$ the matrix $K_S^\theta$ is positive definite implying that $\det(K_S^\theta) > 0$. The determinantal clustering process probability measure (Eq. (2)) therefore places positive mass on every $\mathcal{S} \in \mathbb{S}$. This is a crucial property which is highly attractive in a clustering task as it removes the need to prespecify the number of clusters the data should be partitioned into.

The geometric interpretation of the determinant of an $M \times M$ matrix is the *volume* spanned by its rows. A small determinant implies the rows are 'similar' to each other, whilst conversely a large determinant implies 'dissimilarity' amongst the rows. This phenomenon is illustrated in Figure 1. The reciprocal of the determinant is therefore a natural measure of similarity between data points.

Taking the product of such reciprocal determinants is a simple yet natural way to amalgamate the score for each cluster into a global score for a given partition. It also permits very simple sampling and inference which is discussed later.

Notice that unlike many popular clustering methods e.g. $k$-means or mixtures of Gaussians, the DCP does not assume that each cluster has a mean about which points occur with elliptically symmetrically. Such methods are often highly sensitive to initialisation and outliers which can be avoided by using a DCP type cluster scoring method. We explore this idea further in the Experiments section.

#### 2.1.1 Relation to Gaussian Clustering in Feature Space

Suppose $\phi : \mathcal{X} \to \mathbb{R}^P$ is a non linear feature mapping for some $P \in \mathbb{N}$. Given data $x_1, ..., x_N \in \mathcal{X}$, let $\Phi$ be the $P \times N$ matrix with $n^{th}$ column equal to $\phi(x_n)$. Now imagine that each row of $\Phi$ is an independent draw from a multivariate Gaussian with mean 0 and covariance $\Sigma$. The probability of $\Phi$ is given by

$$L(\Phi|\Sigma) \propto \frac{1}{\sqrt{\det(\Sigma)}} \exp\Big(-\frac{1}{2}\text{Trace}(\Phi \Sigma^{-1} \Phi^\top)\Big). \quad (3)$$

The maximum likelihood estimator of $\Sigma$ is $\hat{\Sigma} = \Phi^\top \Phi$. Note that $\hat{\Sigma}$ is a kernel matrix with $\hat{\Sigma}_{m,n} = \phi(x_m)^\top \phi(x_n)$. Finally it is easy to show that

$$L(\Phi|\hat{\Sigma}) \propto \frac{1}{\sqrt{\det(\Phi^\top \Phi)}} \exp\Big(-\frac{1}{2}\text{Trace}(\Phi(\Phi^\top \Phi)^{-1}\Phi^\top)\Big)$$
$$\propto \det(\Phi^\top \Phi)^{-1/2}. \quad (4)$$

The points $\phi(x_1), ..., \phi(x_N)$ are close together if and only if the covariance between points $\phi(x_1)_p, ..., \phi(x_N)_p$ is small for each $p \in \{1, ..., P\}$. This is the case exactly when $\det(\Phi^\top \Phi)$ is small. The probability of $\Phi$ is therefore large when $\phi(x_1), ..., \phi(x_N)$ are close together.

Suppose we partition the data and fit a Gaussian with the maximum likelihood covariance matrix to each partition. The resulting likelihood would be proportional to a product of terms like in Eq 4. This expression is identical to the DCP likelihood where we simply set $K = \Phi^\top \Phi$ and temper the likelihood (which is discussed in Section 2.2). The DCP does not require explicit specification of the feature map $\phi$ and works entirely on the kernel matrix. Note that clustering the columns of $\Phi$ leads to an entirely different process; Gaussian mixture modelling in feature space [Wang et al., 2003].

#### 2.1.2 A Simple Example

Consider the delta kernel function

$$k(i, j, \theta) = \begin{cases} \theta & \text{if } x_i = x_j, \\ 0 & \text{otherwise.} \end{cases} \quad (5)$$

Recall that $x_i \neq x_j$ for $i \neq j$, therefore for any $A \subseteq X$ of size $N_A$, $K_A^\theta = \theta I_{N_A}$ where $I_{N_A}$ is an $N_A \times N_A$ identity matrix and $\det(K_A^\theta) = \theta^{N_A}$. In fact for any partition $\mathcal{S} \in \mathbb{S}$, $p(\mathcal{S} \in \mathbb{S}|\theta) \propto \theta^N$ which is independent of $\mathcal{S}$. For this choice of kernel, the DCP therefore places *uniform mass over the set of* all partitions. This is what we would expect intuitively since the kernel suggests that points are only similar to themselves and dissimilar to everything else.

#### 2.1.3 Discouraging Singleton Clusters

A potential concern with such a model is that it may favour a large number of very small clusters. A similar problem was observed by Wu and Leahy [1993] when proposing a clustering scheme based on a minimum cut method, where leaf nodes tended to belong to their own clusters. We show that such a drawback does not exist under a DCP framework. A proof of the following can be found in Gentle [2007].

**Lemma 1.** *Suppose $K$, a symmetric positive definite matrix, is written in black matrix form,*

$$K = \begin{pmatrix} A & C^T \\ C & B \end{pmatrix},$$

*then $\det(K) = \det(A)\det(B - CA^{-1}C^T)$.*

For a kernel function $k$ and hyperparameters $\theta \in \Theta$, consider a set $A \subset X$ and $x \in X \backslash A$. By applying Lemma 1, note that

$$\det\left(K_{A \cup \{x\}}^\theta\right)$$
$$= \det\left(K_A^\theta\right) \det\left(K_{\{x\}}^\theta - K_{\{x\},A}^\theta {K_A^\theta}^{-1} K_{A,\{x\}}^\theta\right)$$
$$= \det\left(K_A^\theta\right) \times \left(K_{\{x\}}^\theta - K_{\{x\},A}^\theta {K_A^\theta}^{-1} K_{A,\{x\}}^\theta\right)$$
$$\leq \det\left(K_A^\theta\right) \times \left(K_{\{x\}}^\theta\right)$$
$$= \det\left(K_A^\theta\right) \det\left(K_{\{x\}}^\theta\right),$$

where we use the fact that the determinant of a scalar is the scalar and the inequality comes from the fact that $y{K_A^\theta}^{-1}y^T > 0$ for any non-zero vector $y$ by positive definiteness. We trivially deduce that

$$\det\left(K_{A \cup \{x\}}^\theta\right)^{-1} \geq \det\left(K_A^\theta\right)^{-1} \det\left(K_{\{x\}}^\theta\right)^{-1}$$

and hence that the DCP would always prefer to add a singleton to an existing cluster rather than to assign it to a new one.

It is important to appreciate that such a result does not necessarily hold when comparing the union of two sets each of size greater than 1 i.e. for $A, B \subset X$ disjoint each containing more than 1 element, it may be the case that

$$\det \left(K_{A \cup B}^\theta\right)^{-1} < \det \left(K_A^\theta\right)^{-1} \det \left(K_B^\theta\right)^{-1}.$$

If this were never possible the DCP would be a poor model as, we could show inductively, that its mode would be at the clustering where all points are clustered together for any set of data points $X$.

### 2.1.4 Choosing a Kernel Function

The behaviour of the DCP is entirely encoded in the functional form of the kernel and its parameters. This is entirely analogous to the fact that the behaviour of a support vector machine or a function drawn from a Gaussian process prior with mean 0 is entirely encoded in its covariance kernel function.

Many classes of positive definite kernel functions and more information about Gaussian processes can be found in Rasmussen and Williams [2006]. The squared exponential kernel is a common choice of kernel and is defined by

$$k(x, x', \{l, \sigma\}) = \sigma^2 \exp\left(-\frac{1}{2}(x-x')^\top \mathrm{Diag}(l)^{-1}(x-x')\right), \quad (6)$$

where $\mathrm{Diag}(l)$ is a diagonal matrix with $l_i$ as the $i^{th}$ diagonal entry. For any $N \times N$ positive definite matrix $K$, $\det(\alpha K) = \alpha^N \det(K)$. Since the DCP is a probability measure the constant multiplier becomes redundant hence we can set $\sigma = 1$ without losing any modelling flexibility.

### 2.2 Using a 'Temperature' Parameter

The addition of a *temperature* parameter to the DCP likelihood adds another layer of flexibility to the clustering process. Consider

$$p(\mathcal{S} \in \mathbb{S}|\theta) \propto \prod_{S \in \mathcal{S}} \det \left(K_S^\theta\right)^{-\tau}, \quad (7)$$

where $\tau \in \mathbb{R}_+$. This parameter has the effect of determining how peaked or flat the density is analogous to the temperature parameter in simulated annealing. Here, a large $\tau$ will make the density highly peaked at the mode whilst a small $\tau$ will encourage a uniform density over all partitions.

#### 2.2.1 Kernel Hyperparameters

In some types of data analysis a user may actually know a good, application specific choice of kernel function and hyperparameters they wish to use. In such a case the DCP may be used as a prior over all possible clusterings directly with no further parameter learning required.

In most cases kernel hyperparameters are unknown apriori and have to be learned from data. We proceed under this assumption. In particular, we consider the case of observing many data points only some of which have been labelled. This is developed further in Section 3.1.2.

## 3 SEMI SUPERVISED CLUSTERING WITH DCP

Unlike for a classification problem, the 'name' of a particular cluster is irrelevant. For example, consider the clustering $\{\{1, 4\}, \{2, 3\}\}$. Whether we call the set $\{1, 4\}$ 'cluster 1' or 'cluster 2' is arbitrary, the important information is that 1 and 4 belong to the same cluster whilst 2 and 3 belong to another one.

We therefore can encode the relevant information about the clustering of points in $X$ using a binary indicator matrix $C$, where for $x_i, x_j \in X$

$$C(x_i, x_j) = \begin{cases} 1 & \text{if } x_i,\ x_j \text{ in the same cluster,} \\ 0 & \text{otherwise.} \end{cases} \quad (8)$$

Moreover notice that every partition $\mathcal{S} \in \mathbb{S}$ will have a unique such binary matrix representation which we denote $C_\mathcal{S}$.

In the semi-supervised setting we assume that some portion of our data set is labelled. Let $Z \subset X$ be the set of observed points for which we have labels, i.e. we observe the binary indicator matrix $\hat{C}_Z$ defined on pairs of inputs $x_i, x_j \in Z$.

For a given kernel function and hyperparameters $\theta \in \Theta$ our DCP likelihood function becomes

$$p(\mathcal{S} \in \mathbb{S}|\hat{C}_Z, \theta, \tau) \quad (9)$$
$$\propto \prod_{S \in \mathcal{S}} \det \left(K_S^\theta\right)^{-\tau} \prod_{x,y \in Z} \mathbb{I}(C_\mathcal{S}(x, y) = \hat{C}_Z(x, y)),$$

where $\mathbb{I}(.)$ is an indicator which takes value 1 when its argument is true and 0 otherwise. This second product encodes all the observed labels into the DCP model.

### 3.1 INFERENCE

Assuming a given parameterized kernel function, we describe a Gibbs based sampling method for allocating unlabelled data points to clusters and a MCMC step for learning kernel hyperparameters.

### 3.1.1 Sampling clusters

Suppose the sampler is currently at a particular partition $\mathcal{S} = \{S_1, ..., S_M\}$ for some integer $M \leq N$. Further suppose that for each cluster $S_m$, we have $K^{\theta}_{S_m}{}^{-1}$ stored in memory. We wish to update the cluster location of point $x \in X \backslash Z$ given the clustering of the remaining points. Without loss of generality, suppose $x \in S_M$ and let $\mathcal{S} \backslash \{x\} = \{S_1, ..., S_{M'}\}$ where $M' = M - 1$ if $S_M = \{x\}$ and $M' = M$ otherwise. In the latter case, we remove $x$ from $S_M$ and update $K^{\theta}_{S_M}{}^{-1}$ using the following lemma (proof found in Gentle [2007]).

**Lemma 2.** *Suppose we know $K_A^{-1}$ for some non-empty set $A$. If we add an element $x \in X \backslash A$ to the set $A$, we have*

$$K_{A \cup \{x\}}^{-1} = \begin{pmatrix} U & V \\ V^T & \frac{1}{w} \end{pmatrix},$$

*where*

$$w = K_{\{x\}} - K_{\{x\},A} K_A^{-1} K_{A,\{x\}},$$
$$U = K_A^{-1} + \frac{1}{w} K_A^{-1} K_{A,\{x\}} K_{\{x\},A} K_A^{-1},$$
$$V = -\frac{1}{w} K_A^{-1} K_{A,\{x\}}.$$

We now must assign $x$ to a particular cluster. For $m \in \{1, ..., M'\}$,

$$p(x \in S_m | \mathcal{S} \backslash \{x\}, \theta, \tau) = \frac{p(\{S_1, .., S_m \cup \{x\}, .., S_{M'}\} | \theta, \tau)}{p(\{S_1, ..., S_{M'}\} | \theta, \tau)}$$

$$\propto \frac{\det\left(K^{\theta}_{S_m \cup \{x\}}\right)^{-\tau}}{\det\left(K^{\theta}_{S_m}\right)^{-\tau}}$$

$$\propto \left(K^{\theta}_{\{x\}} - K^{\theta}_{\{x\},S_m} K^{\theta}_{S_m}{}^{-1} K^{\theta}_{S_m,\{x\}}\right)^{-\tau}, \quad (10)$$

and for $m = M' + 1$,

$$p(x \in S_m | \mathcal{S} \backslash \{x\}, \theta, \tau) = \frac{p(\{S_1, ..., S_{M'}, \{x\}\} | \theta, \tau)}{p(\{S_1, ..., S_{M'}\} | \theta, \tau)}$$

$$\propto \det\left(K^{\theta}_{\{x\}}\right)^{-\tau}$$

$$\propto K^{\theta}_{\{x\}}{}^{-\tau}. \quad (11)$$

We therefore allocate $x$ to an existing cluster or a new cluster using a discrete uniform sample with these computed probabilities and update $K^{\theta}_{S_m}{}^{-1}$ using Lemma 2. This procedure is repeated for each $x' \in X \backslash Z$.

Note the conceptual similarity between this sampler and the collapsed Gibbs sampler for Dirichlet process mixtures; for each data point, the sampler decides whether to assign it to an existing (10) or new (11) cluster.

### 3.1.2 Sampling Kernel Hyperparameters and Temperature

Given a particular partition of the data $\mathcal{S} \in \mathbb{S}$, we wish to update the kernel hyperparameters and the temperature parameter using MCMC. We take $\psi = (\theta, \tau)$ to represent all these parameters in the proceeding discussion. Assume a prior density $p(\psi)$ over the parameter space $\Theta \times \mathbb{R}_+$. The posterior density for $\psi$ is given by

$$p(\psi | \mathcal{S}) = \frac{p(\mathcal{S} | \psi) p(\psi)}{\int p(\mathcal{S} | \psi') p(\psi') d\psi'}. \quad (12)$$

We conjecture that the normalising constant of the DCP is analytically intractable. Whilst this is difficult to prove formally, we believe it to be true in part due to the sheer size of the set $\mathbb{S}$, known as the $N^{th}$ Bell number [Wilf, 2006].

Consequently, we say that the posterior density is *doubly intractable* as the integral in the denominator is intractable and the likelihood in the numerator has an intractable normalising constant. A typical Metropolis-Hastings MCMC step would require the ability to compute the numerator of this posterior exactly.

To combat this issue we appeal to the Exchange Sampling algorithm of Murray et al. [2006] where we generate auxiliary data to avoid the need to compute normalising constants for the likelihood. Given that the current hyperparameters are set to $\psi \in \Theta \times \mathbb{R}_+$, suppose we have a proposal distribution $q(\psi \to \psi')$. The Single Variable Exchange Algorithm says to sample $\psi' \sim q(\psi \to \psi')$ and then to sample an auxiliary data set $\mathcal{S}' \sim p(\mathcal{S}' | \psi')$ (note that this can be done using the Gibbs based method of Section 3.1.1). The acceptance probability is set to

$$a = \min\left(1, \frac{q(\psi \to \psi') p(\mathcal{S} | \psi')}{q(\psi' \to \psi) p(\mathcal{S} | \psi)} \times \frac{p(\mathcal{S}' | \psi)}{p(\mathcal{S}' | \psi')}\right). \quad (13)$$

Notice the normalising constants for the observed data likelihoods cancel with those of the auxiliary data likelihoods removing the need to compute them explicitly. Murray et al. [2006] show that using such an acceptance probability, the Markov chain converges to the required posterior in the limit.

## 4 RELATED WORK

It is natural to question the relationship between determinantal clustering and spectral clustering [Shi and Malik, 2000]. Whilst both methods have similar matrix based inputs, their processes are fundamentally different. Spectral clustering maps this similarity matrix to its eigenspace and then performs a simple clustering algorithm e.g. $k$-means. Similarly kernel $k$-means [Dhillon et al.] maps the data to some feature space and performs $k$-means in this new space. In both examples, the kernel matrix is used to map inputs to a latent feature space before performing a simple clustering algorithm. This is not the case for the determinantal clustering process. The DCP simply uses the kernel to ensure positive definiteness so that determinants can be used as a measure of the size of a cluster. The DCP also does not require the prespecification of the number of clusters and learns this from the data.

The Dirichlet Process Gaussian Mixture Model (DPGMM) is a popular nonparametric Bayesian tool for clustering e.g. . This model assumes that each cluster is generated by an independent Gaussian distribution whose parameters are learnt from the data. Such a model requires modelling the joint distribution of all the data which can be difficult in high dimensions. Conversely the DCP requires just the ability to compute a kernel function between pairs of inputs.

There are plenty of flexible discriminative nonparameteric Bayesian models for multi-class classification problems based on transformed Gaussian processes [Lawrence and Jordan, 2005, Sindhwani et al., 2007, Chu et al., 2007, Rogers and Girolami, 2007, Adams and Ghahramani, 2009]. However, these all require knowledge of the number of classes apriori and are inappropriate for clustering tasks where the number of clusters is unknown.

Nonparametric clustering in spectral space is possible using the similarity-dependent Chinese restaurant process [Socher et al., 2011] or by combining the ideas of the DP-means algorithm [Kulis and Jordan, 2012] and kernel $k$-means to get a hard clustering which is able to infer the number of clusters.

## 5 EXPERIMENTS

We implement determinantal clustering on both synthetic and real data sets to demonstrate its properties. To provide a benchmark of results we compare the performance of the DCP with three other popular clustering methods: $k$-means, spectral clustering [Shi and Malik, 2000] and DPGMM [Escobar and West, 1995]. The DPGMM is a generative nonparametric Bayesian model whilst the other two methods require pre-specification of the number of clusters.

Two clustering metrics are computed to reflect the quality of clusterings sampled by these algorithms: adjusted rand index (ARI) Hubert and Arabie [1985] and normalized mutual information (NMI) [Manning et al., 2008]. Both are popular metrics for unsupervised clustering tasks. In cases where classes of data are actually known, we may use classification metrics such as precision and recall, however, we assume that we do not have this knowledge and only are interested in pairwise relationships between points. The ARI takes its maximum value at 1 for a perfect match in clustering, 0 represents a clustering which is equivalent in score to a random clustering and the ARI can also take negative values. The NMI is also maximized at 1 for a perfect clustering, but cannot take negative values. In our experiments, we compare these scores for the unlabelled test data points.

### 5.1 SYNTHETIC DATA

We illustrate two useful properties of the DCP over other clustering methods in this section. One common underlying assumption of clustering models is that data from a particular cluster are distributed elliptically symmetrically about some single cluster mean. Under this paradigm, the further away you are from a cluster mean, the less likely a point is to belong to that cluster. In many instances such an assumption is not a valid one and can lead to poor results. In the synthetic experiments we assume that the number of clusters is known. We use the squared exponential kernel for spectral and determinantal clustering learning the hyperparameters using cross validation and MCMC respectively. When computing clustering metrics for the DCP or DPGMM we sample partitions from the posterior and average scores over samples. For $k$-means and spectral clustering we average scores over a number of alternative initializations.

#### 5.1.1 Clusters with Overlapping Boundaries

Consider the 2 cluster problem illustrated in Figure 2(a). Each cluster was generated from a two-dimensional Gaussian distribution and a few points were added near the boundary of the clusters which are not necessarily closer to their own cluster mean than the other cluster mean. In this experiment all the points other than the ones in squares were given labels and the task was to predict the cluster assignment of these points.

The performance of all models on the synthetic experiments is shown in Table 1. All models other than the DCP struggle with this type of data, especially $k$-means and spectral clustering since both procedures assign points to the clusters whose mean they are closest to. The DPGMM does slightly better as it allocates points probabilistically, but the DCP is the outright winner. This is precisely due to the use of volume spanned by all points as opposed to distance from one point when determining clusters. Moving 1 point over a boundary may severely penalise the squared distance from the mean without affecting the cluster volume as adversarially. In this sense, the DCP is a more robust model.

Table 1: Results of Synthetic Experiments

|         |     | DCP   | DPGMM | $k$-means | Spectral |
|---------|-----|-------|-------|-----------|----------|
| Overlap | ARI | **0.051** | 0.006 | -0.007 | -0.004 |
|         | NMI | **0.143** | 0.062 | 0.044 | 0.046 |
| Multi Modal | ARI | **0.213** | -0.052 | -0.127 | -0.103 |
|         | NMI | **0.382** | 0.176 | 0.150 | 0.122 |

### 5.1.2 Multi-Modal Clusters

Clusters of data points may actually be multimodal in their feature spaces. In such a case choosing a model which assumes elliptical symmetry about a single point is a poor choice. In Figure 2(b) we consider 2 clusters where the first is drawn from a mixture of two Gaussians and the second is drawn from a mixture of 3 Gaussians. Again, all points but the ones in black squares are labelled and the task was to predict the cluster assignments of these points.

Notice that an entire Gaussian mixture in the second cluster is unobserved. From the results in Table 1 we see that DPGMM, $k$-means and spectral clustering all perform poorly. Both DPGMM and $k$-means struggle because of their initialised cluster means. The hidden mixture component is roughly equidistant from these means, so during prediction roughly half of these points are assigned to one cluster and half to the other. Spectral clustering struggles as the learnt kernel parameters are essentially overfit to the training data. The performance of DCP is significantly better than these other models again reflecting the potential benefits of a volume based cluster size metric.

## 5.2 REAL WORLD DATA

### 5.2.1 Wheat Kernels

This data set due to Charytanowicz et al. [2010] is a collection of geometric properties of 3 types of wheat kernels: Kama, Rosa and Canadian. The properties are real valued and include the wheat kernels' area, perimeter, compactness, length, width, asymmetry coefficient and length of kernel groove. We randomly select 20 examples from each type of wheat kernel to construct our data set. In this experiment we observe 6 labelled data points (3 from each of 2 randomly se-lected wheat types) and leave 5 data points from each wheat type as unobserved test points. The remaining 39 points are observed but unlabelled. The task was to predict the cluster assignments of the 15 test points. The results of the experiments are shown in Table 2.

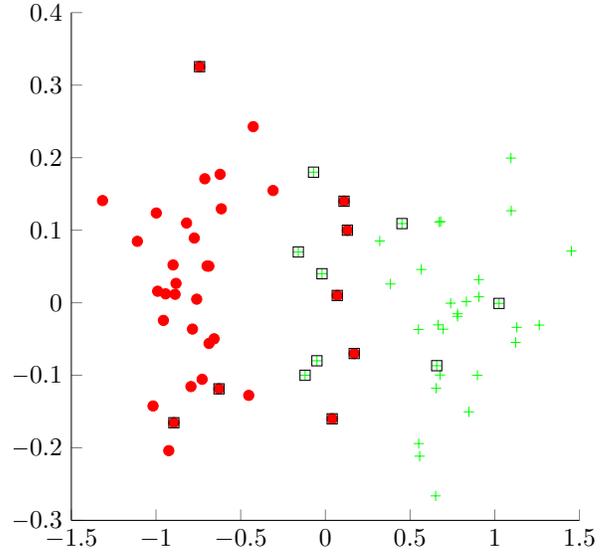

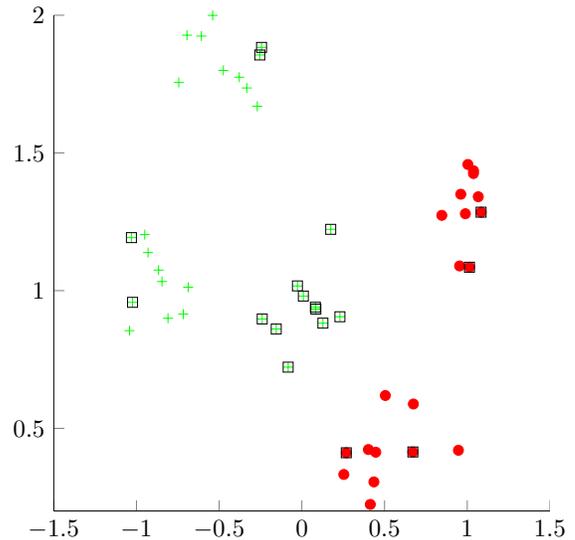

Figure 2: Synthetic datasets. Points to be predicted have dark squares.

In this experiment the DPGMM outperforms other methods. Since the data is 7-dimensional, a mixture of Gaussians is still a powerful technique to use. For $k$-means and spectral clustering, the number of clusters has to be prespecified. Notice that there are only 2 clusters in the training data and that for $k = 2$ both models are poor choices. For $k = 3$, spectral clustering

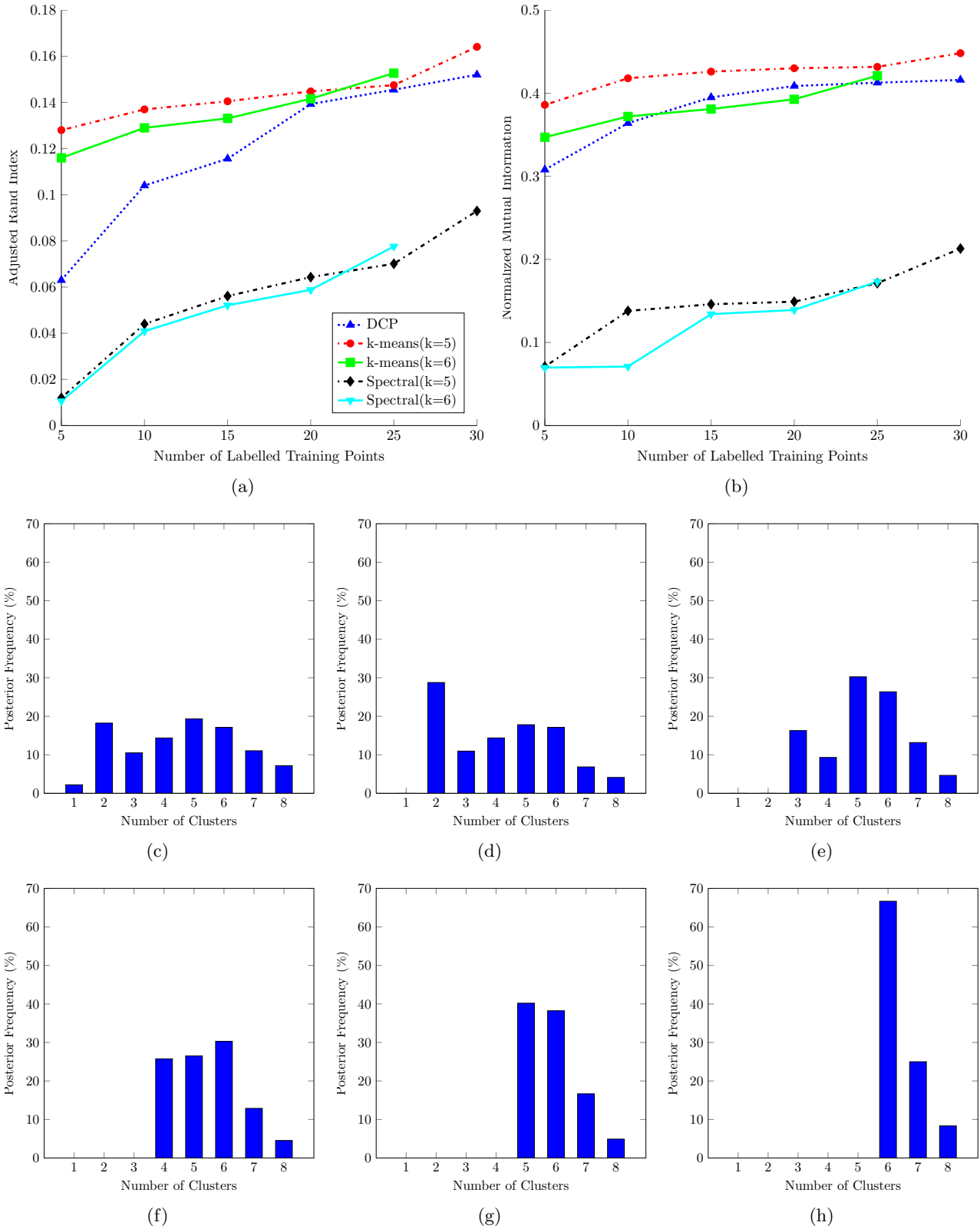

Figure 3: Results of the Car Lane Occupancy data experiments. We show (a) ARI and (b) NMI for each method varying the number of labelled training points. In (c)-(h) we plot the DCP posterior sampled number of clusters for 5, 10, 15, 20, 25 and 30 labelled points respectively.

Table 2: Results of Wheat Kernel Experiments

|     | DCP | DPGMM | k-means k=2 | k-means k=3 | Spectral k=2 | Spectral k=3 |
|-----|-----|-------|-------------|-------------|--------------|--------------|
| ARI | 0.696 | **0.773** | 0.355 | 0.513 | 0.397 | 0.707 |
| NMI | 0.767 | **0.834** | 0.510 | 0.608 | 0.527 | 0.778 |

does well, but only marginally better than the DCP.

#### 5.2.2 Car Lane Occupancy Data

This data was collated by Cuturi [2011] from the Californian Department of Transportation PEMS website. The data describes occupancy rates between 0 and 1 of San Francisco bay area freeways every 10 minutes of every day for 15 months. A total of 963 road detectors were used. Hence for each day we have a $144 \times 963 = 138672$ long feature vector. The task is to cluster data from different days of the week together.

Since this data is extremely high dimensional, using a DPGMM is simply infeasible. In our experiments we extract data points every 2 hours rather than every 10 minutes, leaving our data 11566 dimensional and still beyond the capability of the DPGMM model. However, we are still able to compute a kernel between these feature vectors. In this experiment we consider a 1 parameter squared exponential kernel which has a shared lengthscale parameter across all dimensions.

First we select 6 days of the week: Saturday, Sunday and 4 weekdays. For each day we randomly select 20 data points and set aside 5 from each group as unseen test points; this gives 90 training and 30 test points. We vary the number of labelled points and try to predict the test points. In the $i^{th}$ experiment we assume $5 \times i$ of the training points are labelled and belong in equal numbers to $i$ different clusters. Therefore not only do we vary the number of labels, we also vary the number of observed cluster labels. The results are shown in Figure 3.

Spectral clustering results were generally poor here and this seems to be due to low flexibility of the kernel which only has 1 parameter. Despite the DCP using the same type of kernel function, it has significantly better results. This is due to the increased flexibility offered by the temperature parameter. The posterior sampler seemed to have a mode at around 4 for this data, which suggests that the 1 parameter kernel function was not sufficient in differentiating clusters.

We observe good performance from the $k$-means algorithm, in particular when we set $k$ to the true value of 6. It should be noted that for $k = 5$ however, the DCP has competitive performance versus $k$-means as the number of labels increase, suggesting that when the number of clusters is truly unknown the DCP can be a powerful tool.

The sampled number of clustered under the DCP framework appears slightly multimodal at first. The peak at 2 is due to the process partitioning the weekend against the weekdays. This feature is most pronounced when labels from 2 clusters are observed in Figure 3(b) (one label was for a weekday the other for a weekend day). In Figure 3(f) there is no posterior mass on 1, 2 or 3 clusters and this is because labels for 4 clusters are observed so there is 0 likelihood of the data having less than 4 clusters.

## 6 CONCLUSIONS AND FUTURE WORK

In this work we have presented a novel kernel-based nonparameteric Bayesian approach to learning clusters in data. The key insight involves the use of kernel matrix determinants to score how close together subsets of data points are to each other. We discuss some elegant properties of the process and demonstrate its performance against other popular clustering methods.

Using a volume based cluster measurement proved beneficial for clusters which were not necessarily spread symmetrically about some mean point. A nonparametric Bayesian approach was shown to be fruitful when labelled data was scarce, whilst spectral clustering tended to overfit the kernel hyperparameters in cross validation, especially when the labelled data came from a small number of clusters in relation to the total number of clusters in the data set.

One drawback of such a model is that the computational cost of one cycle of Gibbs updates is $O(N^3)$ since each update requires matrix multiplication which is up to $O(N^2)$. An interesting research direction may be to use existing theory in matrix approximations to improve on this cost. Having a non-analytic normalising constant in the DCP likelihood adds another layer of difficulty in inference; exchange sampling is expensive. In future work it may be worth approximating this constant or using a variational method which makes hyperparameter learning relatively easy.

The remarkable property of this model is the fact that it overcomes the typically difficult task of dealing with complex high dimensional data. As long as any sensible positive definite kernel can be computed between pairs of data points, the determinantal clustering process can infer interesting properties about the data. This feature, combined with not having to prespecify the number of clusters makes the DCP a great contender for analysing complex data sets such as biological sequences, images, text and other multimedia.